\documentclass{easychair}

\usepackage{graphicx} 
\usepackage{amsmath,amssymb,amsfonts}
\usepackage[para]{threeparttable}
\usepackage{booktabs} 
\title{Results of the 2023 CommonRoad Motion Planning Competition for Autonomous Vehicles}

\author{
    Niklas Kochdumper\inst{1}
\and
    Youran Wang\inst{2}
\and
    Johannes Betz\inst{3}
\and
    Matthias Althoff\inst{2}
}

\institute{
   Université Paris Cité, IRIF, 
   Paris, France \\
   \email{niklas.kochdumper@irif.fr}
 \and 
  Technische Universit\"at M\"unchen, 
  Munich, Germany\\
  \email{youran.wang@tum.de,althoff@in.tum.de}
   \and 
  Technische Universit\"at M\"unchen, Autonomous Vehicle Systems Lab,
  Munich, Germany\\
  \email{johannes.betz@tum.de}
}

\authorrunning{}
\date{\phantom{date}\\}
\titlerunning{CommonRoad Motion Planning Competition}

\begin{document}

\maketitle

\begin{abstract}
In recent years, different approaches for motion planning of autonomous vehicles have been proposed that can handle complex traffic situations. However, these approaches are rarely compared on the same set of benchmarks. To address this issue, we present the results of a large-scale motion planning competition for autonomous vehicles based on the CommonRoad benchmark suite. The benchmark scenarios contain highway and urban environments featuring various types of traffic participants, such as passengers, cars, buses, etc. The solutions are evaluated considering efficiency, safety, comfort, and compliance with a selection of traffic rules. This report summarizes the main results of the competition. 
\end{abstract}

\section{Introduction}


The CommonRoad Motion Planning Competition was established in 2021, aiming to bring together researchers working on motion planning for autonomous vehicles. This report summarizes the results from the 2023 edition. 
The main goal of the competition is to provide a fair comparison of different motion planning approaches on a large number of realistic traffic scenarios. To achieve this, all motion planners are executed on the same hardware and consider the same traffic scenarios. Moreover, the benchmark scenarios are realistic since real road networks are used and the behaviors of the traffic participants are either adapted from real-world recordings or simulated by state-of-the-art traffic simulators. Finally, the competition additionally also considers a realistic vehicle model with nonlinear dynamics and parameters taken from a real Ford Escort. 
In addition to presenting the results of the competition, this report also contains a short description of the motion planners submitted by the participants, which provides some insights into the different motion planning strategies used. 

The remainder of this report is organized as follows: First, the format of the competition is described in Sec.~\ref{sec:format} and the rules for performance evaluation are presented in Sec.~\ref{sec: evaluation}. Afterward, a description of the participating motion planners is provided in Sec.~\ref{sec: participants}, before presenting the results of the competition in Sec.~\ref{sec:results}.

\section{Format of the Competition}
\label{sec:format}

\begin{figure}
    \centering
    \includegraphics[width=0.7\linewidth]{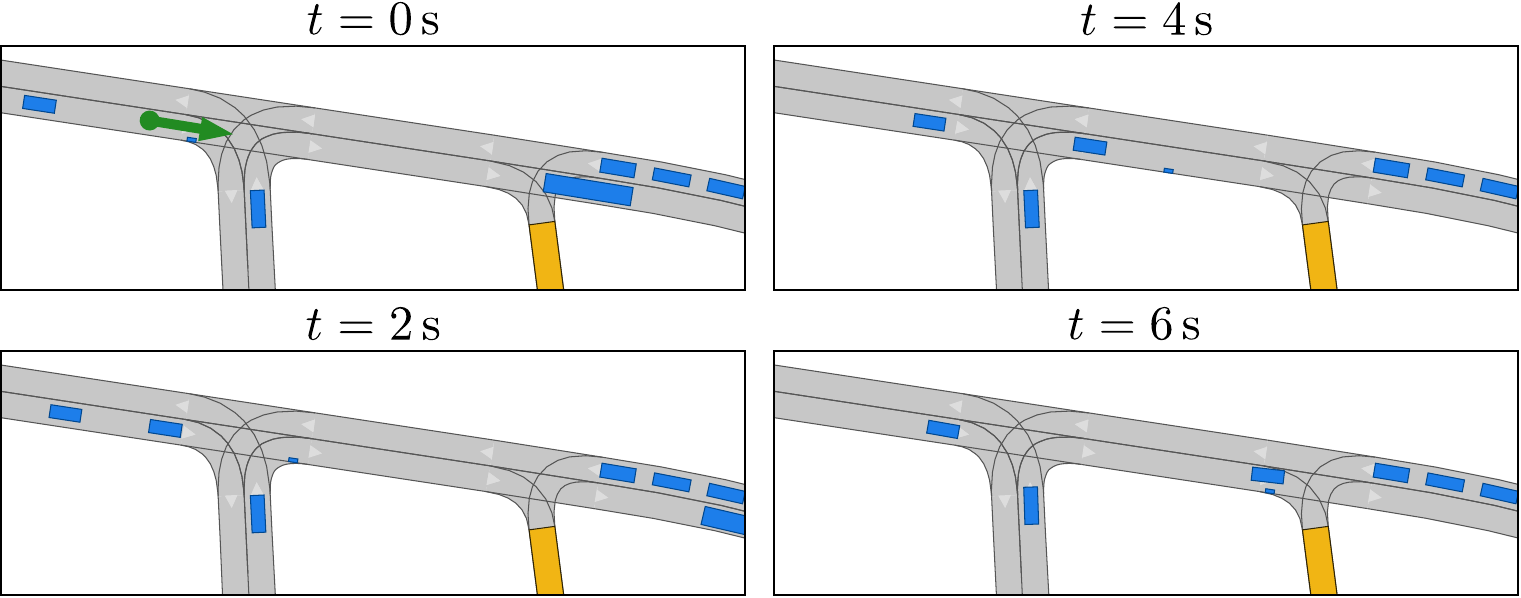}
    \caption{Example for a motion planning problem defined by the CommonRoad scenario \textit{DEU\_Flensburg-73\_1\_T-1}, where the goal set is depicted in yellow, the initial position and velocity of the ego vehicle in green, and the other traffic participants in blue.}
    \label{fig:commonRoadScenario}
\end{figure}

Teams participating in the competition solve motion planning problems for autonomous vehicles; an example is shown in Fig.~\ref{fig:commonRoadScenario}. Traffic scenarios covered in the competition contain highway and urban environments, and feature various types of traffic participants, such as passenger cars, buses, and bicycles. The participants are required to solve motion planning problems encoded in the benchmark scenarios, i.e., develop motion planners to generate feasible trajectories that drive the ego vehicle to the predefined goal region. In the 2023 edition, more than $500$ scenarios have been provided to cover many aspects of autonomous driving.	

The benchmark problems are provided in the CommonRoad format \cite{Althoff2017a}, and the participants upload Docker images with their motion planners to the servers of the CommonRoad website for evaluation. Apart from the number of solved planning problems, submissions are evaluated considering efficiency, safety, and compliance with a selection of traffic rules. The exact evaluation criteria have been decided by an independent jury and are presented in detail in Sec.~\ref{sec: evaluation}. To support teams without a large software framework for motion planning, we provide useful tools for motion planning within the CommonRoad framework, such as a drivability checker \cite{PekIV20}, route planner\footnote{\url{https://gitlab.lrz.de/cps/commonroad-route-planner}}, criticality measurement toolbox \cite{lin2023crime}, etc. The competition provides two types of problems:
	\begin{itemize}
		\item \textbf{Non-interactive} scenarios in which other traffic participants do not react to the behavior of the ego vehicle with provided predictions.
		\item \textbf{Interactive} scenarios using the SUMO \cite{Krajzewicz2012, Klischat2019a} traffic simulator, in which other traffic participants react to the behavior of the ego vehicle.
	\end{itemize}
 In order to lower the entry barrier for the competition, the competition is divided into two phases:
\begin{itemize}
  \item \textbf{Phase I} is for didactic purposes (results are not considered for the evaluation), where the participants can familiarize themselves with the CommonRoad framework. We use public, known scenarios in this phase. The scenarios are a mix of non-interactive and interactive scenarios.
  \item \textbf{Phase II}: The evaluation is solely performed in Phase II. We use non-public, unknown, and interactive scenarios in this phase. The participants provide Docker images of their motion planners. Submitted solutions will appear in the leaderboard\footnote{\url{https://commonroad.in.tum.de/challenges/id/d5a01f69-e828-436d-9f9d-dabf4226e29f}} of the challenge.
\end{itemize}
At the end of the competition, the results were presented in a workshop at the IEEE International Conference on Intelligent Transportation Systems.

\section{Evaluation}
\label{sec: evaluation}
In this section, we present the evaluation criteria of the benchmark solutions, which were decided by the jury chair of the competition. We first verify if a planned trajectory is feasible (see Sec. \ref{sec: feasible_trajectories}) and reaches the goal region of the benchmark scenario. The feasible trajectories that reach the goal region are then evaluated by a cost function (see Sec. \ref{sec: cost function}).
\subsection{Feasible Trajectories}
\label{sec: feasible_trajectories}
A trajectory is classified as feasible if it fulfills the following conditions:
\begin{itemize}
\item \textbf{Collision-free}: The occupancy of the ego vehicle does not intersect with other obstacles within the planning horizon.
\item \textbf{Kinematically feasible}: Trajectories must be feasible regarding the dynamic of a given vehicle model\footnote{\url{https://gitlab.lrz.de/cps/commonroad-vehicle-models}}. In this competition, a kinematic single-track model \cite[Sec.~III.B]{Althoff2017a} configured by the parameters of a real vehicle is employed.
\item \textbf{Road-compliance}: The ego vehicle must stay within the road network and must not occupy walkways and bicycle paths. 
\end{itemize}
We perform the evaluation of the three conditions by the CommonRoad drivability checker \cite{PekIV20}.

\subsection{Cost Function}
\label{sec: cost function}
We evaluate the optimality of feasible trajectories by the following cost function:
\begin{equation}
\label{eq: cost}
    J_\mathrm{ego}=\boldsymbol{w}\,[J^{\mathrm{lon}}_J,~J_\mathrm{SR},~J_\mathrm{D},~J_\mathrm{LC}]^\top,
\end{equation}
where the individual cost terms are defined as follows:
\begin{itemize}
	\item \textbf{Jerk}: $ J^{\mathrm{lon}}_{J} = \int_{t_0}^{t_f} \dddot{s}(t) ^2 \, \mathtt{d}t$, using longitudinal jerk $\dddot{s}(t) $.
	\item \textbf{Steering rate}: $J_\mathrm{SR} = \int_{t_0}^{t_f} v_{\delta}(t)^2 \, \mathtt{d}t$, using steering velocity $v_{\delta}(t)$.
	\item \textbf{Lane center offset}: $J_\mathrm{LC} = \int_{t_0}^{t_f} d(t)^2 \, \mathtt{d}t$, where $d(t)$ is the distance to the lane center.
	\item \textbf{Distance to obstacles}: $J_\mathrm{D} = \int_{t_0}^{t_f} \max(\xi_1, \ldots, \xi_o) \, \mathtt{d}t$, where $o$ is the number of surrounding obstacles in front of the ego vehicle, $\xi_i = e^{-w_\mathtt{dist} d_i}$, $d_i$ is the distance of the ego vehicle to an obstacle, and $w_\mathtt{dist}$ is an additional required weight.
\end{itemize}
The weighting factors are $\boldsymbol{w}=[0.01,~22,~8,~5]$ and $w_\mathtt{dist}=0.2$. The cost function~(\ref{eq: cost}) is implemented in the drivability checker with the ID TR1.
\section{Participants}
\label{sec: participants}
The universities that participated in the competition are subsequently introduced in alphabetical order, where each participant provides a short description of the motion planner that was used for the competition.

\subsection{Stony Brook University}
\label{sec: Stony Brook University}
Niklas Kochdumper and Stanley Bak formed the team of the Reliable Systems Laboratory from the Department of Computer Science at Stony Brook University. This group mainly researches the verification of autonomy, cyber-physical systems, and neural networks. 

The core element of the developed motion planner is a reachability-based decision module \cite{Kochdumper2024}, which uses reachable sets to identify the most suitable driving corridor for the ego vehicle. This module uses the following simplified vehicle model for decision making:
\begin{equation}
	\begin{bmatrix} \xi(t_{i+1}) \\ v(t_{i+1}) \end{bmatrix} = \underbrace{\begin{bmatrix} 1 & \Delta t \\ 0 & 1\end{bmatrix}}_{A} \begin{bmatrix} \xi(t_i) \\ v(t_i) \end{bmatrix} + \underbrace{\begin{bmatrix} 0.5 \, \Delta t^2 \\ \Delta t \end{bmatrix}}_{B} a(t_i),
    \label{eq:doubleIntegrator}
\end{equation}
where $\xi(t)$ is the longitudinal position of the vehicle along the lanelet \cite{Bender2014}, $v(t)$ is the velocity, $a(t)$ is the acceleration, $\Delta t$ is the time step size, and $t_i = i \cdot \Delta t$ are the time points for time-discretization. Since the simplified vehicle model \eqref{eq:doubleIntegrator} is linear and only has two states, the reachable set $\mathcal{R}(t)$ for this model can be computed very efficiently using polygons as a set representation:
\begin{equation*}
		\mathcal{R}(t_{i+1}) = A \, \mathcal{R}(t_i) + B \, [-a_{\text{max}},a_{\text{max}}],
\end{equation*}
where $a_{\text{max}}$ is the maximum acceleration of the vehicle.

\begin{figure}[!tb]
    \centering
    \includegraphics[width=\linewidth]{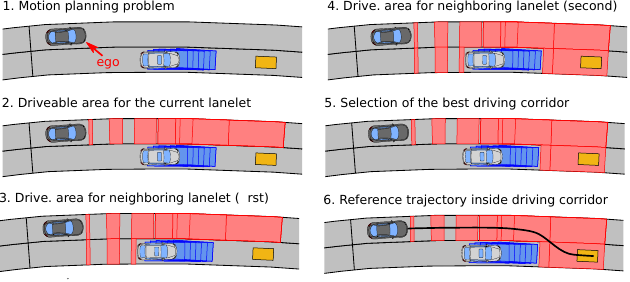}
    \caption{Visualization of the single steps for the reachability-based decision module \cite{Kochdumper2024} used by the motion planner from Stony Brook University, where the goal set is depicted in yellow, the space occupied by other traffic participants in blue, the drivable area for the ego vehicle in red, and the provided reference trajectory in black.}
    \label{fig:decisionModule}
\end{figure}

The overall process for constructing all possible driving corridors is visualized in Fig.~\ref{fig:decisionModule}: The procedure starts by computing the reachable set for the current lanelet. Next, the computed reachable set is used to check if any neighboring or successor lanelets are reachable. If this is the case, the reachable sets for all reachable lanelets are computed, and it is again checked if any additional lanelets are reachable. This procedure is repeated until all possible driving corridors have been identified. Finally, the most suitable driving corridor is selected from all driving corridors reaching the goal set based on a cost function that penalizes the number of lane changes as well as deviations from a desired velocity profile. Improvements and extensions for this basic procedure consist in applying criteria based on the friction circle to ensure that the selected driving corridor is also driveable by the real vehicle, and considering traffic rules like speed limits by removing regions from the driving corridor that violate the rules.


In addition to the most suitable driving corridor, the decision module \cite{Kochdumper2024} described above also provides a reference trajectory inside this corridor. However, since this reference trajectory is constructed using the simplified vehicle model in \eqref{eq:doubleIntegrator}, it might not be driveable by the nonlinear kinematic single track model $\dot x = f(x,u)$ with state vector $x \in \mathbb{R}^5$ and control input vector $u \in \mathbb{R}^2$ used in the CommonRoad competition. Therefore, we solve the following optimal control problem to obtain a trajectory that is consistent with the dynamics of the nonlinear vehicle model and close to the reference trajectory $z_{\text{ref}}(t)$ provided by the decision module:
\begin{equation}
\begin{split}
    & \quad \quad \quad \quad \min_{u_1,\dots,u_N} ~ \sum_{i=1}^N (C \, x(t_i) - z_{\text{ref}}(t_i))^T Q \, (C \, x(t_i) - z_{\text{ref}}(t_i)) + u_i^T \, R \, u_i \\
    & \text{subject to} \\
    & \quad \quad \quad \quad x(t_{i+1}) = x(t_i) + \int_0^{\Delta t} f(x(t_i),u_i) \, dt, \quad \quad i = 0,\dots N-1, \\
    & \quad \quad \quad \quad -u_{\text{max}} \leq u_i \leq u_{\text{max}} \quad \quad i = 0,\dots N-1,
\end{split}
\label{eq:optimalControl}
\end{equation}
where $N$ is the number of discrete time steps, $Q \in \mathbb{R}^{2 \times 2}$ and $R \in \mathbb{R}^{2 \times 2}$ are user-defined weighting matrices, $u_{\text{max}} \in \mathbb{R}^2$ is the limit for the control inputs, and $C \in \mathbb{R}^{2 \times 5}$ is a matrix that selects the x- and y-position of the vehicle from the state vector $x(t)$ of the vehicle model.

For the implementation of the motion planner, we used the code for the decision module \cite{Kochdumper2024} available on GitHub\footnote{\url{https://github.com/KochdumperNiklas/MotionPlanner}} and applied CasADi\footnote{\url{https://web.casadi.org/}} to solve the optimal control problem \eqref{eq:optimalControl}. Moreover, the parameter values we used are $\Delta t = 0.1$s, $Q = I_2$, $R = 0.01 \cdot I_2$, where $I_n \in \mathbb{R}^{n \times n}$ denotes the identity matrix.

\subsection{Technical University of Munich}
\label{sec: Technical University of Munich}
Our motion planner FRENETIX \cite{trauth2024FRENETIX} was developed by the Autonomous Vehicle Systems (AVS) Laboratory from the Technical University of Munich. The research of the AVS Lab focuses on the development of new algorithms that enable trajectory and behavior planning, adaptive control, and continuous learning of the systems. The team participating in the competition consists of Alexander Hobmaier, Rainer Trauth, and Johannes Betz.

FRENETIX offers a high-performance, modular solution that integrates various aspects of motion planning, ensuring comfort, safety, and precision in complex urban scenarios. A significant contribution of FRENETIX is its modular design, which realizes easy adaptability and scalability, making it possible to integrate new features or updates without requiring a complete overhaul of the system. Moreover, the FRENETIX algorithm is implemented in both Python and C++, offering flexibility while ensuring real-time capability.

\begin{figure}
    \centering
    \includegraphics[width=1\linewidth]{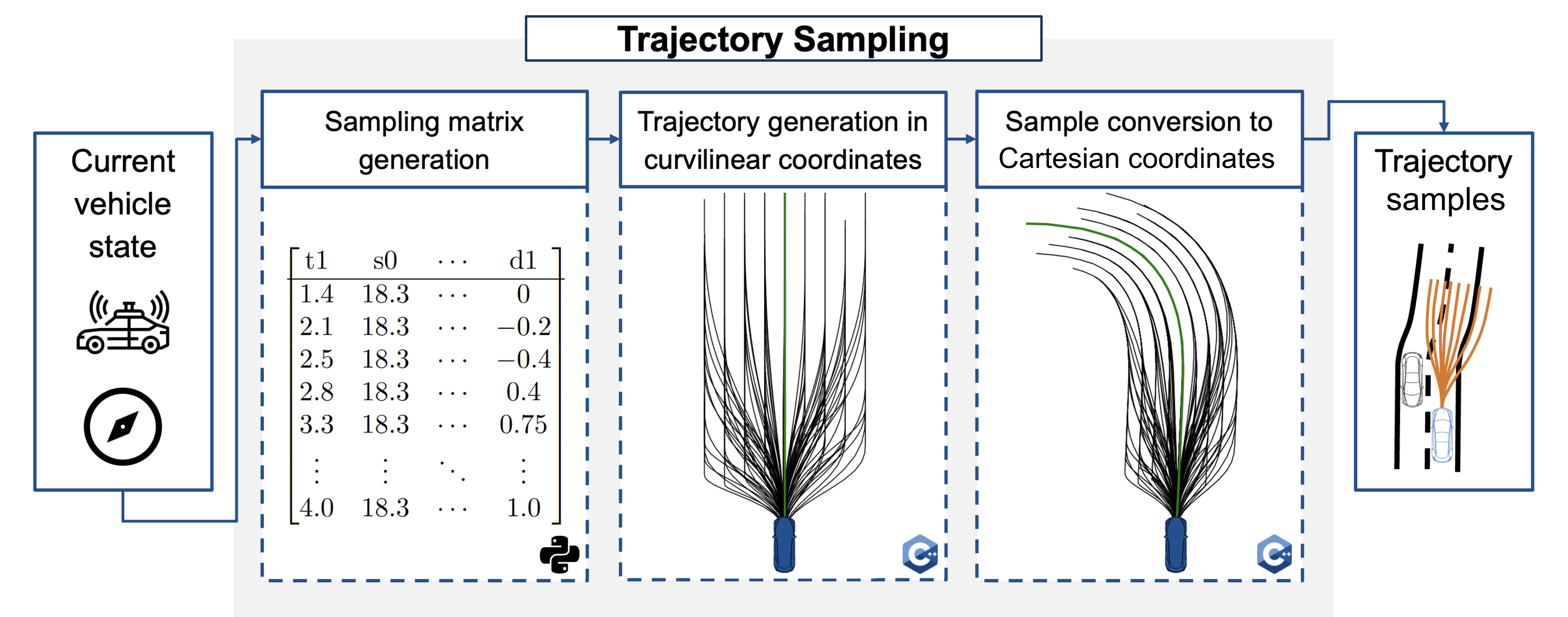}
    \includegraphics[width=1\linewidth]{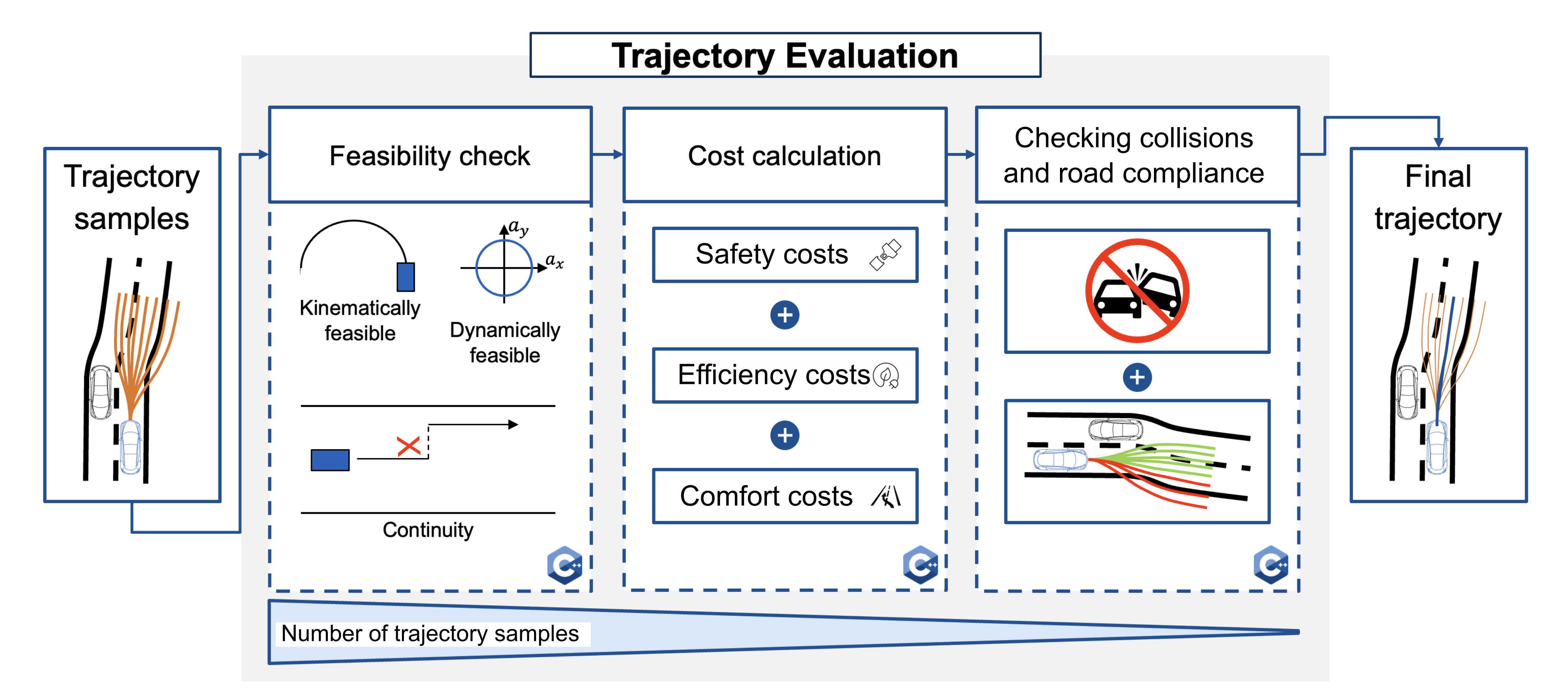}
    \caption{The individual steps during trajectory sampling and trajectory evaluation in the FRENETIX motion planner.}
    \label{fig:sampling-Evaluation}
\end{figure}

\subsubsection{FRENETIX Architecture}

FRENETIX is based on a sampling-based method, which is effective in high-dimensional spaces, searching for free space and target states. The first step in the planning process is to calculate the reference path through global planning, setting the stage for more detailed planning at a local level. The motion planning cycle in FRENETIX involves several critical steps:

\begin{enumerate}
    \item \textbf{Trajectory sampling:} Trajectory sampling (Figure~\ref{fig:sampling-Evaluation}) is the first component of the FRENETIX planning process. Potential final states of the ego vehicle are generated based on a predetermined discretization scheme as in \cite[Table 1]{trauth2024FRENETIX}. By connecting the current state of the ego vehicle and the potential terminal states with polynomial functions, multiple trajectory samples are produced. These samples are initially represented in curvilinear coordinates, which align with the road geometry and realize efficient sampling. Subsequently, the samples are converted to Cartesian coordinates, facilitating their use in trajectory evaluation and further processing.

    \item \textbf{Trajectory evaluation:} In the second component, the sampled trajectories undergo various trajectory evaluation steps (Figure \ref{fig:sampling-Evaluation}) to determine their feasibility. These include kinematic and dynamic feasibility checks, as well as collision avoidance and road compliance checks. Additionally, the trajectories are evaluated based on cost functions that account for efficiency, safety, and comfort.

    \item \textbf{Scenario update and iteration:} After evaluation, the best trajectory is selected. If the goal has not been reached, the scenario is updated, and the planning cycle iterates, refining the trajectory until the vehicle successfully reaches its destination.
\end{enumerate}

\begin{figure}
    \centering
    \includegraphics[width=1.0\linewidth]{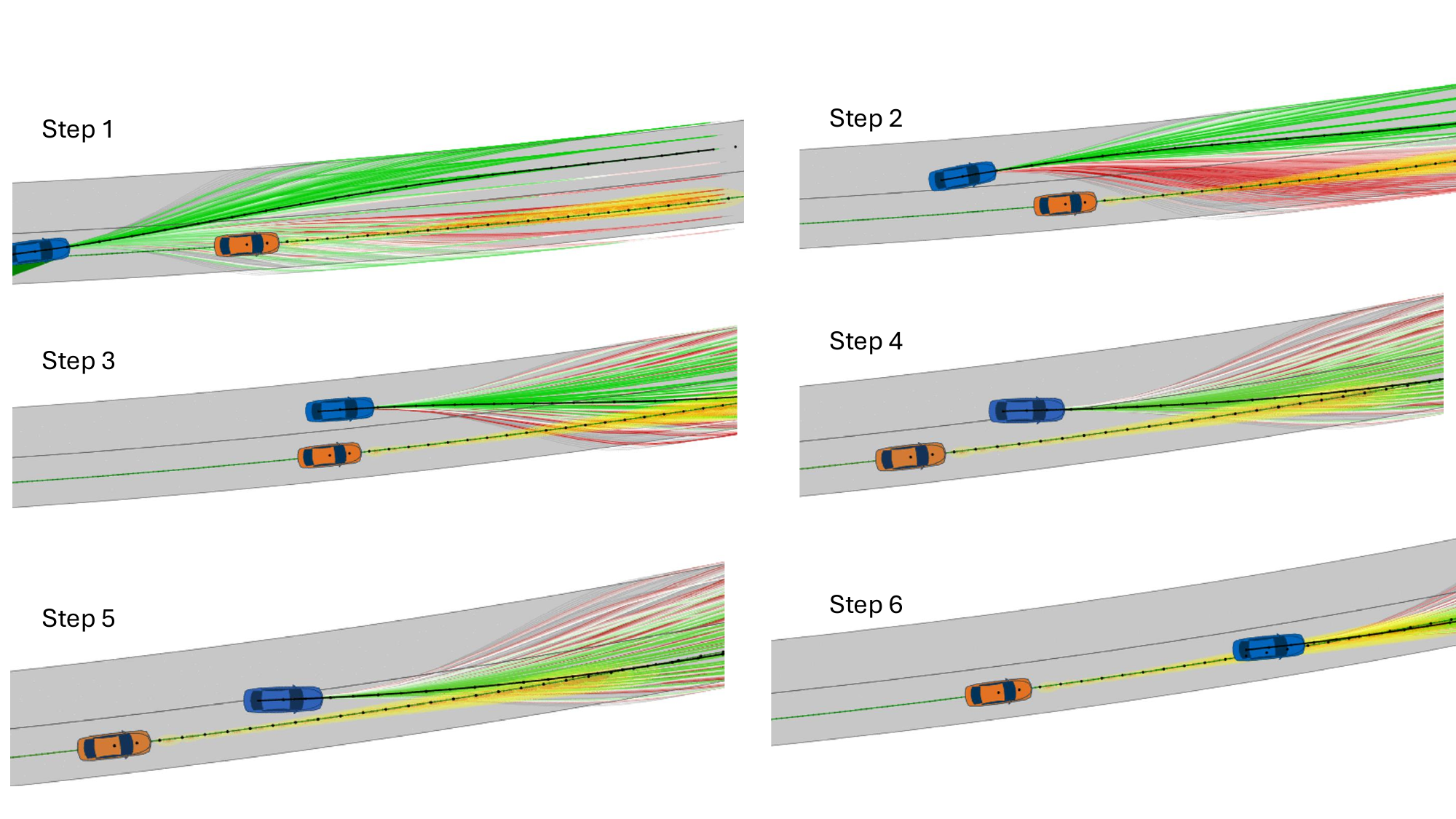}
    \caption{One of the scenarios in which FRENETIX was tested involves dynamic overtaking, a complex maneuver that requires precise trajectory planning and real-time adjustments. FRENETIX successfully navigated this scenario, demonstrating its capability to handle high-stakes driving situations where safety and accuracy are paramount.}
    \label{fig:overtaking}
\end{figure}
\subsubsection{FRENTIX Performance}
The effectiveness of the FRENETIX motion planner is demonstrated by its performance in various scenarios and its runtime efficiency. Besides the competition, FRENETIX was tested on 1,750 scenarios from the CommonRoad database. FRENETIX successfully solved 1,539 scenarios, representing an 88\% success rate. The remaining scenarios were either infeasible, resulted in collisions, or exceeded the time limit. These results highlight the robustness of the planner in handling a wide range of driving situations (Figure \ref{fig:overtaking}).

The runtime analysis provides insight into the computational efficiency of FRENETIX, which was evaluated using both single-core and multi-core implementations in C++ and Python. The results indicate that the C++ implementation, particularly in a multi-core setup, significantly outperforms the Python version, especially as the number of trajectories increases. For instance, when processing 90,000 trajectories, the multi-core C++ implementation achieved a runtime of 717 ms, compared to 10,986 ms for the Python multi-core implementation.
\subsubsection{FRENETIX Enhancements with Modules}
The modular architecture of FRENETIX makes it possible to integrate additional modules that enhance its functionality. Among these, the motion prediction module \cite{WALENET} plays a pivotal role by enabling the planner to anticipate the movements of other vehicles and objects in the environment, thereby improving its decision-making process in dynamic and fast-changing scenarios. Equally important is the occlusion-aware planning module \cite{trauth2023toward}, which enables FRENETIX to consider hidden areas that might conceal obstacles, further increasing the safety and reliability of its planned trajectories. The architecture also supports reinforcement learning optimization, allowing the system to learn from past experiences and refine its decision-making algorithms. Coupled with an additional risk assessment, these advancements empower FRENETIX to tackle increasingly complex and unpredictable driving environments. 

\begin{table*}[t]
    \caption{Results of the motion planners introduced in Sec. \ref{sec: participants}.}
    \vspace{2pt}
    \label{tab: results}
    \centering
    \begin{threeparttable}
        \renewcommand{\arraystretch}{1.1}
        \begin{tabular}{@{}lcc@{}} 
            \toprule \vspace{2pt}
            \textbf{Participants} & \textbf{Solved scenarios} & \textbf{Top $1$ solutions$^*$}\\
            \midrule 
            \vspace{2pt} Technical University of Munich (Sec.~\ref{sec: Technical University of Munich}) & 130 & 84\\[2pt]
            Stony Brook University (Sec. \ref{sec: Stony Brook University}) & 116 & 54\\[2pt]
            \bottomrule
        \end{tabular}
        \begin{tablenotes}
            \item[*] Multiple submitted planners might be able to solve a particular scenario. The solution with the lowest cost according to function~(\ref{eq: cost}) is referred to as the top 1 solution to the scenario. 
        \end{tablenotes}
    \end{threeparttable}
\end{table*}

\section{Results}
\label{sec:results}

The submitted motion planners are evaluated on $230$ interactive CommonRoad benchmark scenarios that are unknow to the participants. The evaluation is timed out after 6 hours, and at most two cores can be used. Note that the time limit aims at penalizing computation time, since motion planners that are faster will be able to solve more scenarios within the given time frame. All computations are carried out on a server with two AMD EPYC 7763 processors with $2$ TB memory.

For each submission in Sec. \ref{sec: participants}, the number of solved scenarios and the number of best solutions are presented in Tab. \ref{tab: results}. Both motion planners were able to successfully solve a large number of scenarios within the given time frame. Consequently, both planners overall have a quite similar performance, and the victory in the competition was very close.
The winner of the competition is the team of Alexander Hobmaier, Rainer Trauth, and Johannes Betz from Technical University of Munich, whose planner results in the most solutions with the lowest costs. Niklas Kochdumper and Stanley Bak from Stony Brook University won the second place in the competition.

\begin{figure}
    \centering
    \includegraphics[width=1\linewidth,trim={0.4cm 0.5cm 0 0},clip]{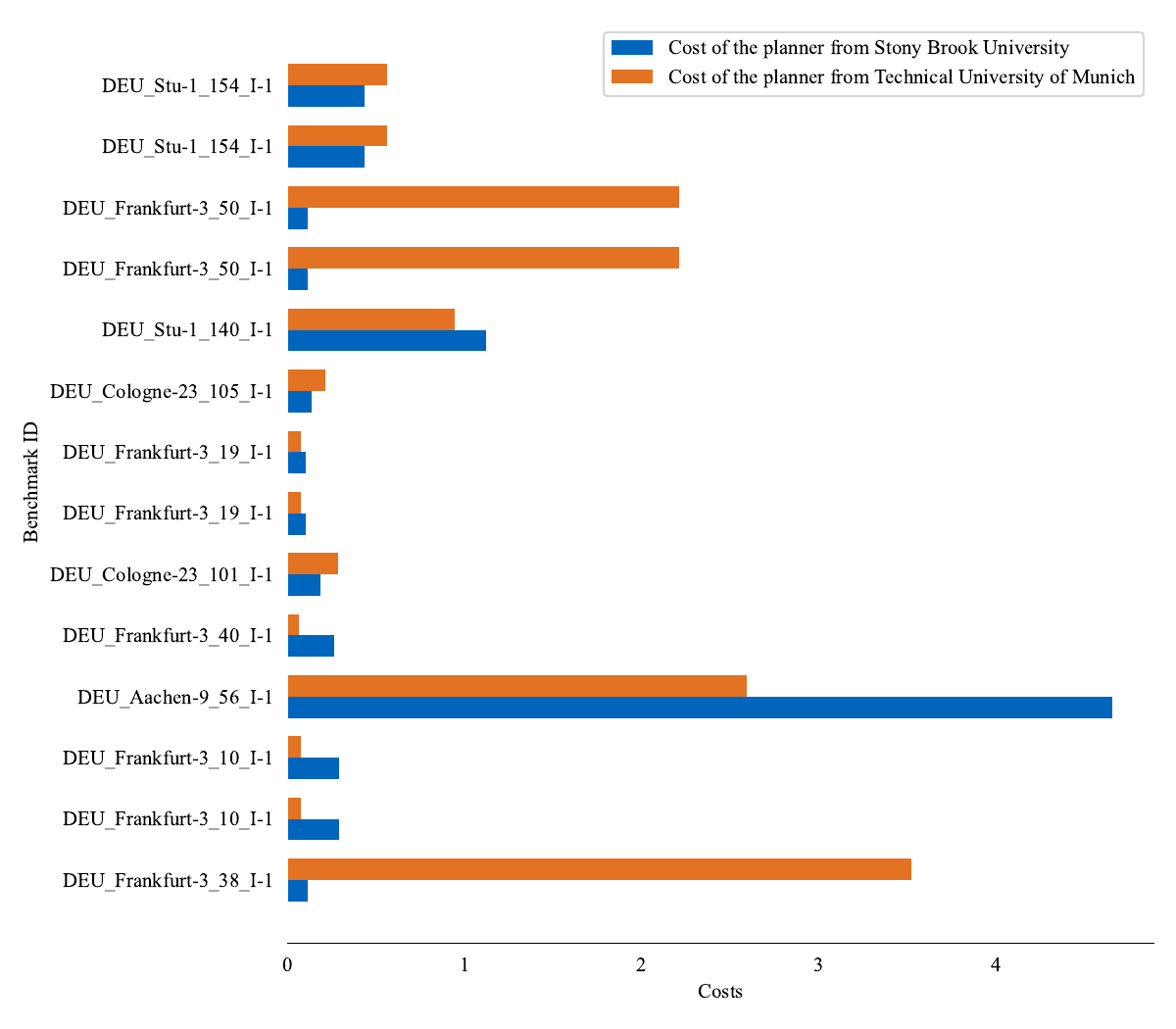}
    \caption{Performance comparison on benchmark scenarios that are solved by both planners.}
    \label{fig: cost}
\end{figure}

For a more detailed comparison of the two participating motion planners, Fig.~\ref{fig: cost} displays the value of the cost function in \eqref{eq: cost} for some exemplary scenarios that could be successfully solved by both motion planners. In many scenarios, both planners provide solutions with very similar and quite low overall costs. Exceptions are the scenarios \textit{DEU\_Frankfurt-3\_50\_I-1}, \textit{DEU\_Frankfurt-3\_50\_I-1}, and \textit{DEU\_Frankfurt-3\_38\_I-1}, where the motion planner from Stony Brook University performs significantly better than the one from Technical University of Munich, as well as the scenarios \textit{DEU\_Frankfurt-3\_40\_I-1}, \textit{DEU\_Aachen-9\_56\_I-1}, \textit{DEU\_Frankfurt-3\_10\_I-1}, and \textit{DEU\_Frankfurt-3\_10\_I-1}, where the motion planner from Technical University of Munich performs much better than the one from Stony Brook University. The results shown in Fig.~\ref{fig: cost} also demonstrate the different degrees of difficulty for the scenarios considered in the competition, since some scenarios are easily solvable by both planners with very low cost, while other scenarios can only be solved with very high cost by both participants.

\section{Conclusion}

This report summarizes the results of the 3\textsuperscript{rd} CommonRoad Motion Planning Competition for Autonomous Vehicles held in 2023. 
Among the participants, two groups from Stony Brook University and Technical University of Munich were awarded for submitting high-performance motion planners which both were able to successfully solve a large number of scenarios. Interestingly, the planners from these two groups apply very different strategies for motion planning, which makes the performance comparison very exciting: While the motion planner from Stony Brook University combines a high-level decision making module with an optimization-based trajectory planner, the motion planner from Technical University of Munich is sampling-based. Despite the very different motion planning strategies used by the participants, their motion planners result in a very similar overall performance in the competition.

\bibliographystyle{abbrv}
\bibliography{bib/general.bib}

\end{document}